# The Computational Complexity of Sensitivity Analysis and Parameter Tuning

**Johan Kwisthout and Linda C. van der Gaag**
Department of Information and Computing Sciences, University of Utrecht
P.O. Box 80.089, 3508 TB Utrecht, The Netherlands
{johank,linda}@cs.uu.nl

## Abstract

While known algorithms for sensitivity analysis and parameter tuning in probabilistic networks have a running time that is exponential in the size of the network, the exact computational complexity of these problems has not been established as yet. In this paper we study several variants of the tuning problem and show that these problems are NP$^{PP}$-complete in general. We further show that the problems remain NP-complete or PP-complete, for a number of restricted variants. These complexity results provide insight in whether or not recent achievements in sensitivity analysis and tuning can be extended to more general, practicable methods.

## 1 Introduction

The sensitivity of the output of a probabilistic network to small changes in the network's parameters, has been studied by various researchers [1, 14, 7, 2, 19, 4]. Whether the parameter probabilities of a network are assessed by domain experts or estimated from data, they inevitably include some inaccuracies. In a sensitivity analysis of the network, the parameter probabilities are varied within a plausible range and the effect of the variation is studied on the output computed from the network, be it a posterior probability or the most likely value of an output variable.

The results of a sensitivity analysis are used, for example, to establish the robustness of the network's output. The results are used also upon engineering a probabilistic network, for example to distinguish between parameters which allow some imprecision and parameters which should be determined as accurately as possible [6]. Another use is for carefully *tuning* the parameter probabilities of a network to arrive at some desired model behavior [3].

Research efforts in sensitivity analysis and parameter tuning for probabilistic networks have resulted in a variety of fundamental insights and computational methods. While the majority of these insights and methods pertain to a one-way sensitivity analysis in which the effect of varying a single parameter probability on a single output probability or output value is studied, recently there also has been some pioneering work on extending these insights to higher-order analyses [2, 6].

The currently available algorithms for sensitivity analysis and parameter tuning of probabilistic networks have a running time that is exponential in the size of a network. This observation suggests that these problems are intractable in general. The actual computational complexity of the problems has not been studied yet, however. In this paper we define several variants of the tuning problem for probabilistic networks and show that these variants are NP$^{PP}$-complete in general. We further show that the tuning problem remains NP-comlete, even if the topological structure of the network under study is restricted to a polytree, and PP-complete, even if the number of conditional probability tables involved is bounded.

Given the unfavorable complexity results obtained, even for restricted cases of the tuning problem, we have that we cannot expect to arrive at efficient, more general computational methods for sensitivity analysis and parameter tuning for probabilistic networks. Our complexity results in fact suggest that further research should concentrate on tuning a limited number of parameters, in networks where inference is tractable.

The paper is organized as follows. After briefly reviewing the basic concepts involved in sensitivity analysis and parameter tuning in Section 2, we present some preliminaries from complexity theory in Section 3 and formally define several variants of the tuning problem in Section 4. We give a general completeness proof for these problems in Section 5. We further address some special, restricted cases of these problems in Section 6. The paper ends with our concluding observations in Section 7.

## 2 Sensitivity analysis and tuning

A probabilistic network $\mathbf{B} = (\mathbf{G}, \Gamma)$ includes a directed acyclic graph $\mathbf{G} = (\mathbf{V}, \mathbf{A})$, where $\mathbf{V} = \{V_1, \ldots, V_n\}$ models a set of stochastic variables and $\mathbf{A}$ models the (in)dependences between them, and a set of parameter probabilities $\Gamma$, capturing the strengths of the relationships between the variables. The network models a joint probability distribution $\Pr(\mathbf{V}) = \prod_{i=1}^{n} \Pr(v_i \mid \pi(V_i))$ over its variables, where $\pi(V)$ denotes the parents of $V$ in $\mathbf{G}$. We will use $\Pr(C = c \mid \mathbf{E} = \mathbf{e})$ to denote the probability of the value $c$ of the output variable $C$, given an instantiation $\mathbf{e}$ to the set of evidence variables $\mathbf{E}$, which will be abbreviated as $\Pr(c \mid \mathbf{e})$. We will denote a particular set of parameter probabilities as $\mathbf{X} \subseteq \Gamma$, and we will use $X$ to denote a single parameter. We will use $\mathbf{x}$ and $x$ to denote the *combination of values* of a set of parameters, respectively the value of a single parameter. In sensitivity analysis and parameter tuning, we are interested in the effect of changes in the parameter probabilities $\mathbf{X}$ on an output probability for a designated variable $C$. The *sensitivity function* $f_{\Pr(c|\mathbf{e})}(\mathbf{X})$ expresses the probability of the output in terms of the parameter set $\mathbf{X}$. We will omit the subscript if no ambiguity can occur.

In a one-way sensitivity analysis, we measure the sensitivity of an output probability of interest with respect to a single parameter. The parameter under consideration is systematically varied from 0 to 1 and the other parameters from the same CPT are co-varied such that their mutual proportional relationship is kept constant [20]. Thus, if the parameter $X = \Pr(b_i \mid \rho)$ (denoting the conditional probability of the value $b_i$ of the variable $B$ given a particular configuration $\rho$ of $B$'s parents) is varied from 0 to 1, the other parameters $Pr(b_j \mid \rho)$ for the variable $B$ are varied such that

$$\Pr(b_j \mid \rho)(X) = \Pr(b_j \mid \rho) \cdot \frac{1 - X}{1 - \Pr(b_i \mid \rho)}$$

for any value $b_j$ other than $b_i$. Under the condition of covariation, the sensitivity function $f(X)$ is a quotient of two linear functions [7] and takes the form

$$f(X) = \frac{c_1 \cdot X + c_2}{c_3 \cdot X + c_4}$$

where the constants can be calculated from the other parameter probabilities in the network.

A one-way sensitivity analysis can be extended to measure the effect of the simultaneous variation of *two* parameters on the output [5]. The sensitivity function then generalizes to

$$f(X_1, X_2) = \frac{c_1 \cdot X_1 \cdot X_2 + c_2 \cdot X_1 + c_3 \cdot X_2 + c_4}{c_5 \cdot X_1 \cdot X_2 + c_6 \cdot X_1 + c_7 \cdot X_2 + c_8}$$

In this function, the terms $c_1 \cdot X_1 \cdot X_2$ and $c_5 \cdot X_1 \cdot X_2$ capture the interaction effect of the parameters on the output variable. This can further be generalized to *n-way* sensitivity analyses [6, 2] where multiple parameters are varied simultaneously. While higher-order analyses can reveal synergistic effects of variation, the results are often difficult to interpret [20].

For performing a one-way sensitivity analysis, efficient algorithms are available that build upon the observation that for establishing the sensitivity of an output probability it suffices to determine the constants in the associated sensitivity function. The simplest method for this purpose is to compute, from the network, the probability of interest for up to three values for the parameter under study; using the functional form of the function to be established, a system of linear equations is obtained, which is subsequently solved [7]. For the network computations involved, any standard propagation algorithm can be used. A more efficient method determines the required constants by propagating information through a junction tree, similar to the standard junction-tree propagation algorithm [12]. This method requires a very small number of inward and outward propagations in the tree to determine either the constants of all sensitivity functions that relate the probability of interest to any one of the network parameters, or to determine the sensitivity functions for any output probability in terms of a single parameter. Both algorithms are exponential in the size of the network, yet have a polynomial running time for networks of bounded treewidth.

Closely related to *analyzing* the effect of variation of parameters on the output—and often the next step after performing such an analysis—is *tuning* the parameters, such that the output has the desired properties. The output may need to satisfy particular constraints, e.g. $\Pr(c \mid \mathbf{e}) \geq q$, $\Pr(c_1 \mid \mathbf{e}) / \Pr(c_2 \mid \mathbf{e}) \geq q$ or $\Pr(c_1 \mid \mathbf{e}) - \Pr(c_2 \mid \mathbf{e}) \geq q$, for a particular value $q$. There are a number of algorithms to determine the solution space for a set of parameters given such constraints [2]. The computational complexity of these algorithms is always exponential in the treewidth $w$ of the graph (i.e., the size of the largest clique in the join-tree), yet varies from $O(c^w)$ for single parameter tuning, to $O(n \cdot \prod_{i=1}^{k} F(X_i) \cdot c^w)$ for tuning $n$ parameters, where $c$ is a constant, $k$ is the number of CPTs that include at least one of the parameters being varied, and $F(X_i)$ denotes the size of the $i$-th CPT. Note that the tuning problem is related to the inference problem in so-called *credal networks* [8], where each variable is associated with sets of probability measures, rather than single values as in Bayesian networks. This problem has been proven $\mathsf{NP}^{\mathsf{PP}}$-complete [9].

Often, we want to select a combination of values for the parameters that satisfies the constraints on the output probability of interest, but has minimal impact on the other probabilities computed from the network. In other cases, we want the modification to be as small as possible. In other words, we want to find a tuning that not merely satisfies the constraints, but is also *optimal*, either with respect to the minimal amount of parameter change needed, or the minimal change in the joint probability distribution induced by the parameter change. Here we discuss two typical distance measures between joint probability distributions, namely those proposed by Kullback and Leibler [13], and Chan and Darwiche [3].

The distance measure introduced by Chan and Darwiche [3], denoted by $D_{CD}$, between two joint probability distributions $\Pr_\mathbf{x}$ and $\Pr_{\mathbf{x}'}$ is defined as:

$$D_{CD}(\Pr_\mathbf{x}, \Pr_{\mathbf{x}'}) \stackrel{def}{=} \ln \max_\omega \frac{\Pr_\mathbf{x}(\omega)}{\Pr_{\mathbf{x}'}(\omega)} - \ln \min_\omega \frac{\Pr_\mathbf{x}(\omega)}{\Pr_{\mathbf{x}'}(\omega)}$$

where $\omega$ is taken to range over the joint probabilities of the variables in the network. The Kullback-Leibler measure [13], denoted by $D_{KL}$, is defined as:

$$D_{KL}(\Pr_\mathbf{x}, \Pr_{\mathbf{x}'}) \stackrel{def}{=} \sum_\omega \Pr_\mathbf{x}(\omega) \ln \frac{\Pr_\mathbf{x}(\omega)}{\Pr_{\mathbf{x}'}(\omega)}$$

Calculating either distance between two distributions is intractable in general. It can be proven that calculating $D_{CD}$ is NP-complete and that calculating $D_{KL}$ is PP-complete[1]. The *Euclidean distance* is a convenient way to measure the *amount of change* needed in $\mathbf{x}$ to go from $\Pr_\mathbf{x}$ to $\Pr_{\mathbf{x}'}$. This distance, denoted by $D_E$, is defined as:

$$D_E(\mathbf{x}, \mathbf{x}') \stackrel{def}{=} \sqrt{\sum_{x_i \in \mathbf{x}, x'_i \in \mathbf{x}'} (x_i - x'_i)^2}$$

The Euclidean distance depends only on the parameters that are changed and can be calculated in $O(|\mathbf{X}|)$.

## 3 Complexity theory

In the remainder, we assume that the reader is familiar with basic concepts of computational complexity theory, such as the classes P and NP, and completeness proofs. For a thorough introduction to these subjects we refer to textbooks like [10] and [16]. In addition to these basic concepts, we use the complexity class PP (Probabilistic Polynomial time). This class contains languages $L$ accepted in polynomial time by a *Probabilistic Turing Machine*. Such a machine augments the more traditional non-deterministic Turing Machine with a probability distribution associated with each state transition, e.g. by providing the machine with a tape, randomly filled with symbols [11]. If all choice points are binary and the probability of each transition is $\frac{1}{2}$, then the *majority* of the computation paths accept a string $s$ if and only if $s \in L$.

A typical problem in PP (in fact PP-complete) is the INFERENCE problem [15, 18]: given a network $\mathbf{B}$, a variable $V_1$ in $\mathbf{V}$, and a rational number $0 \leq q \leq 1$, determine whether $\Pr(V_1 = v_1) \geq q$. Recall that $\Pr(V_1, \ldots, V_n) = \prod_{i=1}^n \Pr(V_i \mid \pi(V_i))$. To determine whether $\Pr(v_1) \geq q$, we sum over all marginal probabilities $\Pr(V_1, \ldots, V_n)$ that are consistent with $v_1$. This can be done using a Probabilistic Turing Machine in polynomial time. The machine calculates the multiplication of conditional probabilities $\Pr(V_i \mid \pi(V_i))$, $i = 1, \ldots, n$, choosing a computation path in which each variable $V_i$ is assigned a value according to the conditional probability $\Pr(V_i \mid \pi(V_i))$. Each computation path corresponds to a specific joint value assignment, and the probability of arriving in a particular state corresponds with the probability of that assignment. At the end of this computation path, we accept with probability $\frac{1}{2} + (\frac{1}{q} - 1)\epsilon$, if the joint value assignment to $V_1, \ldots, V_n$ is consistent with $v_1$, and we accept with probability $(\frac{1}{2} - \epsilon)$ if the joint value assignment is *not* consistent with $v_1$. The majority of the computation paths (i.e., $\frac{1}{2} + \epsilon$) then arrives in an accepting state if and only if $\Pr(v_1) \geq q$.

Another concept from complexity theory that we will use in this paper is *oracle access*. A Turing Machine $\mathcal{M}$ has oracle access to languages in the class A, denoted as $\mathcal{M}^A$, if it can query the oracle in one state transition, i.e., in $O(1)$. We can regard the oracle as a 'black box' that can answer membership queries in constant time. For example, NP$^{PP}$ is defined as the class of languages which are decidable in polynomial time on a non-deterministic Turing Machine with access to an oracle deciding problems in PP. Informally, computational problems related to probabilistic networks that are in NP$^{PP}$ typically combine some sort of *selecting* with *probabilistic inference*.

Not all real numbers are exactly computable in finite time. Since using real numbers may obscure the true complexity of the problems under consideration, we assume that all parameter probabilities in our network are rational numbers, thus ensuring that all calculated probabilities are rational numbers as well. This is a realistic assumption, since the probabilities are normally either assessed by domain experts or estimated by a learning algorithm from data instances. For similar reasons, we assume that $\ln(x)$ is approximated within a finite precision, polynomial in the binary representation of $x$.

---

[1]These results are not yet published but will be substantiated in a forthcoming paper.

## 4 Problem definitions

In the previous sections, we have encountered a number of computational problems related to sensitivity analysis and parameter tuning. To prove hardness results, we will first define decision problems related to these questions. Because of the formulation in terms of decision problems, all problems are in fact tuning problems.

PARAMETER TUNING
**Instance:** Let $\mathbf{B} = (\mathbf{G}, \Gamma)$ be a Bayesian network where $\Gamma$ is composed of rational probabilities, and let Pr be its joint probability distribution. Let $\mathbf{X} \subseteq \Gamma$ be a set of parameters in the network, let $C$ denote the output variable, and $c$ a particular value of $C$. Furthermore, let $\mathbf{E}$ denote a set of evidence variables with joint value assignment $\mathbf{e}$, and let $0 \leq q \leq 1$.
**Question:** Is there a combination of values $\mathbf{x}$ for the parameters in $\mathbf{X}$ such that $\Pr_\mathbf{x}(c \mid \mathbf{e}) \geq q$?

PARAMETER TUNING RANGE
**Instance:** As in PARAMETER TUNING.
**Question:** Are there combinations of values $\mathbf{x}$ and $\mathbf{x}'$ for the parameters in $\mathbf{X}$ such that
$\Pr_\mathbf{x}(c \mid \mathbf{e}) - \Pr_{\mathbf{x}'}(c \mid \mathbf{e}) \geq q$?

EVIDENCE PARAMETER TUNING RANGE
**Instance:** As in PARAMETER TUNING; furthermore let $\mathbf{e_1}$ and $\mathbf{e_2}$ denote two particular joint value assignments to the set of evidence variables $\mathbf{E}$.
**Question:** Is there a combination of values $\mathbf{x}$ for the parameters in $\mathbf{X}$ such that
$\Pr_\mathbf{x}(c \mid \mathbf{e_1}) - \Pr_\mathbf{x}(c \mid \mathbf{e_2}) \geq q$?

MINIMAL PARAMETER TUNING RANGE
**Instance:** As in PARAMETER TUNING; furthermore let $r \in \mathbb{Q}^+$.
**Question:** Are there combinations of values $\mathbf{x}$ and $\mathbf{x}'$ for the parameters in $\mathbf{X}$ such that $D_E(\mathbf{x}, \mathbf{x}') \leq r$ and such that $\Pr_\mathbf{x}(c \mid \mathbf{e}) - \Pr_{\mathbf{x}'}(c \mid \mathbf{e}) \geq q$?

MINIMAL CHANGE PARAMETER TUNING RANGE
**Instance:** As in PARAMETER TUNING; furthermore let $s \in \mathbb{Q}^+$, and let $D$ denote a distance measure for two joint probability distributions as reviewed in Section 2 .
**Question:** Are there combinations of values $\mathbf{x}$ and $\mathbf{x}'$ for the parameters in $\mathbf{X}$ such that $D(\mathbf{x}, \mathbf{x}') \leq s$ and $\Pr_\mathbf{x}(c \mid \mathbf{e}) - \Pr_{\mathbf{x}'}(c \mid \mathbf{e}) \geq q$?

MODE TUNING
**Instance:** As in PARAMETER TUNING; furthermore let $\top(\Pr(C))$ denote the mode of $\Pr(C)$.
**Question:** Are there combinations of values $\mathbf{x}$ and $\mathbf{x}'$ for the parameters in $\mathbf{X}$ such that $\top(\Pr_\mathbf{x}(C \mid \mathbf{e})) \neq \top(\Pr_{\mathbf{x}'}(C \mid \mathbf{e}))$?

Furthermore, we define EVIDENCE MODE TUNING, MINIMAL PARAMETER MODE TUNING, and MINIMAL CHANGE MODE TUNING corresponding to the PARAMETER TUNING variants of these problems.

## 5 Completeness results

We will construct a hardness proof for the PARAMETER TUNING RANGE problem. Hardness of the other problems can be derived with minimal changes to the proof construction. More specifically, we prove NP$^{\text{PP}}$-hardness of the PARAMETER TUNING RANGE-problem by a reduction from E-MAJSAT; this latter problem has been proven complete by Wagner [21] for the class NP$^{\text{PP}}$. We will use a reduction technique, similar to the technique used by Park and Darwiche [17] to prove NP$^{\text{PP}}$-hardness of the PARTIAL MAP-problem.

We first observe that all tuning problems from Section 4 are in NP$^{\text{PP}}$: given $\mathbf{x}$, $\mathbf{x}'$, $q$, $r$ and $s$, we can verify all claims in polynomial time using a PP oracle, since inference is PP-complete [18]. For example, with the use of the oracle, we can verify in polynomial time whether $\Pr_\mathbf{x}(c \mid \mathbf{e}) - \Pr_{\mathbf{x}'}(c \mid \mathbf{e}) \geq q$, for a given $\mathbf{x}$, $\mathbf{x}'$, and $q$. Likewise, we can calculate the Euclidean distance of $\mathbf{x}$ and $\mathbf{x}'$ in polynomial time and verify that it is less than $r$. Determining whether a distance between two joint probability distributions is smaller than $s$ is NP-complete (for the distance $D_{CD}$ defined by Chan and Darwiche [3]) or PP-complete (for the distance $D_{KL}$ defined by Kullback and Leibler [13]). Thus, we can non-deterministically compute an assignment to $\mathbf{X}$ and check (using a PP oracle) that the distance is smaller than $s$. Therefore, all problems are in NP$^{\text{PP}}$.

To prove hardness, we will reduce PARAMETER TUNING RANGE from E-MAJSAT, defined as follows:

E-MAJSAT
**Instance:** Let $\phi$ be a Boolean formula with $n$ variables $V_i$ ($1 \leq 1 \leq n$), grouped into two disjoint sets $\mathbf{V_E} = V_1, \ldots, V_k$ and $\mathbf{V_M} = V_{k+1}, \ldots, V_n$.
**Question:** Is there an instantiation to $\mathbf{V_E}$ such that for at least half of the instantiations to $\mathbf{V_M}$, $\phi$ is satisfied?

We construct a probabilistic network $\mathbf{B}_\phi$ from a given Boolean formula $\phi$ with $n$ variables $V_i$ and instantiation templates $\mathbf{V_E}$ and $\mathbf{V_M}$. For all variables $V_i$, in the formula $\phi$, we create a matching stochastic variable $V_i$ in $\mathbf{V}$ for the network $\mathbf{B}_\phi$, with possible values *true* and *false* with uniform distribution. These variables are roots in the network $\mathbf{B}_\phi$. We denote $X_i = \Pr(V_i = \text{true})$ as the *parameter* of $V_i$.

For each logical operator in $\phi$, we create an additional stochastic variable in the network, whose parents are the corresponding sub-formulas (or single variable in case of a negation operator) and whose conditional probability table is equal to the truth table of that operator. For example, the $\wedge$-operator would have a conditional probability $\Pr(\wedge = \textit{true}) = 1$ if and only if both its parents have the value *true*, and 0 otherwise. We denote the stochastical variable that is associated with the top-level operator in $\phi$ with $V_\phi$. Furthermore, we add a variable $S$ with values *true* and *false*, with uniform probability distribution where $X_S = \Pr(S = \textit{true})$ is the parameter of $S$. Lastly, we have an output variable $C$, with parents $S$ and $V_\phi$ and values *true* and *false*, whose CPT is equal to the truth table of the $\wedge$-operator. The set of parameters $\mathbf{X}$ in the PARAMETER TUNING RANGE problem now is defined to be $\{X_1, \ldots, X_k\} \cup X_S$, i.e., the parameters of the variables in $\mathbf{V_E}$ and the parameter of $S$. We set $q = \frac{1}{2}$.

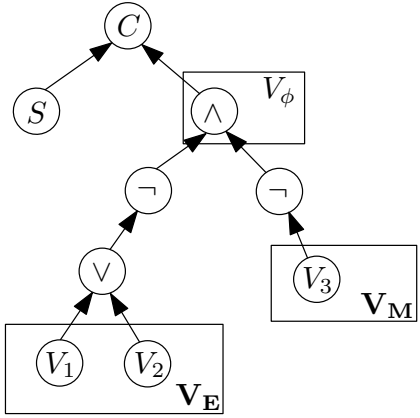

Figure 1: Example of construction

Figure 1 shows the graphical structure of the probabilistic network constructed for the E-MAJSAT instance $(\phi, \mathbf{V_E}, \mathbf{V_M})$, where $\phi = \neg(V_1 \vee V_2) \wedge \neg V_3$, $\mathbf{V_E} = \{V_1, V_2\}$, and $\mathbf{V_M} = \{V_3\}$. Note that this E-MAJSAT instance is satisfiable with $V_1 = V_2 = F$; for that instantiation to $\mathbf{V_E}$, at least half of the possible instantiations to $\mathbf{V_M}$ will satisfy the formula.

**Theorem 1.** PARAMETER TUNING RANGE *is* NP$^{PP}$-*complete.*

*Proof.* Membership of NP$^{PP}$ can be proved as follows. Given $\mathbf{x}'$ and $\mathbf{x}$, we can verify whether $\Pr_{\mathbf{x}}(c \mid \mathbf{e}) - \Pr_{\mathbf{x}'}(c \mid \mathbf{e}) \geq q$ in polynomial time, given an oracle that decides INFERENCE. Since INFERENCE is PP-complete, this proves membership of NP$^{PP}$.

To prove hardness, we construct a transformation from the E-MAJSAT problem. Let $(\phi, \mathbf{V_E}, \mathbf{V_M})$ be an instance of E-MAJSAT, and let $\mathbf{B}_\phi$ be a probabilistic network, with parameters $\mathbf{x} = \{X_1, \ldots, X_k\} \cup X_S$, constructed as described above. Trivially, there exists a combination of parameter values $\mathbf{x}'$ such that $\Pr_{\mathbf{x}'}(C = \textit{true}) = 0$, namely all assignments in which $X_S = 0$. In that case, at least one of the parents of $C$ has the value *false* with probability 1 and thus $\Pr_{\mathbf{x}'}(C = \textit{true}) = 0$.

On the other hand, if $\mathbf{x}$ includes $X_S = 1$, then $\Pr_{\mathbf{x}}(C = \textit{true})$ depends on the values of $X_1, \ldots, X_k$. More in particular, there exist parameter values such that $\Pr_{\mathbf{x}}(C = \textit{true}) \geq \frac{1}{2}$, if and only if $(\phi, \mathbf{V_E}, \mathbf{V_M})$ has a solution. We can construct a solution $\mathbf{x}$ by assigning 1 to $X_S$, 1 to all variables in $\{X_1, \ldots, X_k\}$ where the corresponding variable in $\mathbf{V_E}$ is set to *true*, and 0 where it is set to *false*. On the other hand, if $(\phi, \mathbf{V_E}, \mathbf{V_M})$ is *not* satisfiable, then $\Pr_{\mathbf{x}}(C = \textit{true})$ will be less than $\frac{1}{2}$ for *any* parameter setting. Due to the nature of the CPTs of the 'operator' variables which mimic the truth tables of the operators, $\Pr_{\mathbf{x}}(C = \textit{true}) = 1$ for a value assignment to the parameters that is consistent with a satisfying truth assignment to $\phi$. If there does not exist a truth assignment to the variables in $\mathbf{V_E}$ such that the majority of the truth assignments to the variables in $\mathbf{V_M}$ satisfies $\phi$, then there cannot be a value assignment to $\mathbf{X}$ such that $\Pr_{\mathbf{x}}(C = \textit{true}) \geq \frac{1}{2}$. Thus, if we can decide whether there exist two sets of parameter settings $\mathbf{x}$ and $\mathbf{x}'$ such that in this network $\mathbf{B}_\phi$, $\Pr_{\mathbf{x}}(C = \textit{true}) - \Pr_{\mathbf{x}'}(C = \textit{true}) \geq q$, then we can answer $(\phi, \mathbf{V_E}, \mathbf{V_M})$ as well. This reduces E-MAJSAT to TUNING PARAMETER RANGE. □

Note that the constructed proof shows, that the PARAMETER TUNING RANGE problem remains NP$^{PP}$-complete, even if we restrict the set of parameters to constitute only prior probabilities, if all variables are binary, if all nodes have indegree at most 2, if the output is a singleton variable, and if there is no evidence. We will now show completeness proofs of the other problems.

**Corollary 2.** *All tuning problems defined in Section 4 are* NP$^{PP}$-*complete.*

*Proof.* We will show how the above construction can be adjusted to prove hardness for these problems.

- PARAMETER TUNING: From the above construct, leave out the nodes $S$ and $C$, such that $\mathbf{x} = \{X_1, \ldots, X_k\}$. There is an instantiation $\mathbf{x}$ such that $\Pr_{\mathbf{x}}(V_\phi = \textit{true}) \geq \frac{1}{2}$, if and only if $(\phi, \mathbf{V_E}, \mathbf{V_M})$ has a solution.

- EVIDENCE PARAMETER TUNING RANGE: From the above construct, replace $S$ with a singleton evidence variable $E$ with values *true* and *false* and uniform distribution; denote $E = \textit{true}$ as $e_1$ and $E = \textit{false}$ as $e_2$ and let $\mathbf{x} = \{X_1, \ldots, X_k\}$.

$\Pr_{\mathbf{x}}(C = true \mid E = e_2) = 0$ for all possible parameter settings of $\mathbf{x}$. On the other hand, $\Pr_{\mathbf{x}}(V_\phi = true) \geq \frac{1}{2}$ and thus $\Pr_{\mathbf{x}}(C = true \mid E = e_1) \geq \frac{1}{2}$ if and only if $(\phi, \mathbf{V_E}, \mathbf{V_M})$ has a solution.

- MINIMAL PARAMETER TUNING RANGE and MINIMAL CHANGE PARAMETER TUNING RANGE: These problems have TUNING PARAMETER RANGE as a special case (set $r, s = \infty$) and thus hardness follows by restriction.

- MODE TUNING: Since $C$ has two values, $\Pr(C = false) = 1 - \Pr(C = true)$. In particular, $\top(C) = true$ if $\Pr(C = true) \geq \frac{1}{2}$, and $\top(C) = false$ if $\Pr(C = false) \geq \frac{1}{2}$. If $X_S = 0$ then $\top(C) = false$. $\Pr(C = true) \geq \frac{1}{2}$, if and only if $\top(C) = true$.

EVIDENCE MODE TUNING, MINIMAL PARAMETER MODE TUNING, and MINIMAL CHANGE MODE TUNING: Apply similar construct modifications as with the corresponding PARAMETER TUNING problems. □

## 6 Restricted problem variants

In the previous section, we have shown that in the general case, PARAMETER TUNING RANGE is NP$^{PP}$-complete. In this section, the complexity of the problem is studied for restricted classes of instances. More in particular, we will discuss tuning problems in networks with bounded topologies and tuning problems with a bounded number of CPTs containing parameters to be tuned.

### 6.1 Bounded topologies

In this section we will show that restrictions on the topology of the network alone will not suffice to make the problem tractable. In fact, PARAMETER TUNING RANGE remains hard, even if $\mathbf{B}$ is a polytree. Similar results can be derived for the other problems. To prove NP-completeness of PARAMETER TUNING RANGE on polytrees, we reduce MAXSAT to PARAMETER TUNING RANGE on polytrees, using a slightly modified proof from [17]. The (unweighted) MAXSAT problem is defined as follows:

MAXSAT
**Instance:** Let $\phi$ be a Boolean formula in CNF format, let $\mathbf{C}_\phi = C_1 \ldots C_m$ denote its clauses and $\mathbf{V}_\phi = V_1 \ldots V_n$ its variables, and let $1 \leq k \leq m$.
**Question:** Is there an assignment to the variables in $\phi$, such that at least $k$ clauses are satisfied?

We will construct a polytree network $\mathbf{B}$ as follows. For each variable in the formula, we create a variable in the network with values *true* and *false*, with uniform probability distribution. We denote the parameter of $V_i$ as $X_i$ as in the previous construct. We define a *clause selector variable* $S_0$ with values $c_1, \ldots, c_m$ and uniform probability, i.e. $\Pr(S_0 = c_i) = \frac{1}{m}$. Furthermore, we define *clause satisfaction variables* $S_i$, with values $c_0, \ldots, c_m$, associated with each variable. Every variable $S_i$ has $V_i$ and $S_{i-1}$ as parents. Lastly, we define a variable $V_S$, with values *true* and *false*, with uniform probability distribution, and parameter $X_S$, and a variable $C$ with values *true* and *false*, parents $X_S$ and $S_n$. See Figure 2 for the topology of this network. The CPT for $S_i (i \geq 1)$ and $C$ is given in Table 1. In this table, $T(V_i, j)$ and $F(V_i, j)$ are Boolean predicates that evaluate to 1 if the truth assignment to $V_i$ satisfies, respectively does not satisfy, the $j$-th clause.

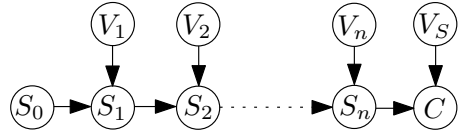

Figure 2: Construction with polytrees

| | $\Pr(S_i \mid V_i, S_{i-1})$ | |
|---|---|---|
| $S_{i-1}$ | $S_i = c_0$ | $S_i \neq c_0$ |
| $c_0$ | 1 | 0 |
| $c_j$ | $T(V_i, j)$ | $F(V_i, j)$ |

| | $\Pr(C = T \mid V_S, S_n)$ | |
|---|---|---|
| $S_n$ | $x_S$ | $\neg x_S$ |
| $c_0$ | 1 | 0 |
| $c_j$ | 1 | 1 |

Table 1: CPT for $\Pr(S_i \mid V_i, S_{i-1})$ and $\Pr(C \mid S_n, V_S)$

**Theorem 3.** PARAMETER TUNING RANGE *remains* NP-*complete if* $\mathbf{B}$ *is restricted to polytrees.*

*Proof.* Membership of NP is immediate, since we can decide INFERENCE in polynomial time on polytrees. Given $\mathbf{x}'$ and $\mathbf{x}$, we can thus verify whether $\Pr_{\mathbf{x}'}(C = c) - \Pr_{\mathbf{x}}(C = c) \geq q$ in polynomial time.

To prove NP-hardness, we reduce MAXSAT to PARAMETER TUNING RANGE. Let $(\phi, k)$ be an instance of MAXSAT. From the clauses $\mathbf{C}_\phi$ and variables $\mathbf{V}_\phi$, we construct $\mathbf{B}_\phi$ as discussed above. Similarly as in the previous proof, if $X_S = 0$ then $\Pr(C = true) = 0$ for any instantiation to the parameters $X_1$ to $X_n$. If $X_S = 1$ then we observe the following. For every instantiation $c_j$ to $S_0$, the probability distribution of $S_i$ is as follows. $\Pr(S_i = c_0 \mid V_i) = 1$ if the instantiation to $V_1 \ldots V_i$ satisfies clause $c_j$, and 0 otherwise. $\Pr(S_i = c_j \mid V_i) = 1$ if this instantiation does *not* satisfy clause $c_j$.

$$\Pr(S_i \mid x_i) = \begin{cases} S_i = c_0, V_{1...i} \text{ satisfies } c_j & 1 \\ S_i = c_j, V_{1...i} \text{ satisfies } c_j & 0 \\ S_i = c_0, V_{1...i} \text{ does not satisfy } c_j & 0 \\ S_i = c_j, V_{1...i} \text{ does not satisfy } c_j & 1 \\ \text{otherwise} & 0 \end{cases}$$

Of course, $\Pr(S_i)$ is conditioned on $V_i$ and thus depends on $X_i$. For $X_i = 0$ or $X_i = 1$, either $\Pr(S_i = c_0) = 1$ or $\Pr(S_i = c_j) = 1$, for intermediate values of $X_i$ the probability mass is shared between $\Pr(S_i = c_0)$ and $\Pr(S_i = c_j)$. But then $\Pr(S_n = c_0)$ is 1 for a particular clause selection $c_j$ in $S_0$, if and only if the parameter setting to $X_1$ to $X_n$ satisfies that clause. Due to the conditional probability table of $C$ and $X_S = 1$, $\Pr_\mathbf{x}(C = true) = 1$ if and only if the parameter setting $\mathbf{x}$ satisfies that clause. Summing over $S_0$ yields $\Pr_\mathbf{x}(C = true) = \frac{k}{n}$, where $k$ is the number of clauses that is satisfied by $\mathbf{x}$. Thus, a PARAMETER TUNING RANGE query with values 0 and $\frac{k}{n}$ would solve the MAXSAT problem. This proves NP-hardness of PARAMETER TUNING RANGE on polytrees. □

### 6.2 Bounded number of CPTs

In the previous section we have shown that a restriction on the topology of the network in itself does not suffice to make parameter tuning tractable. In this section we will show that bounding the number of CPTs containing parameters in $\mathbf{X}$ in itself is not sufficient either. Note that trivial solutions to the PARAMETER TUNING may exist for particular subsets $\mathbf{X}$ of the set of parameter probabilities $\Gamma$. For example, if $\mathbf{X}$ constitutes all conditional probabilities $\Pr(C = c \mid \pi(C))$, for all configurations of parents of $C$, then a trivial solution would set all these parameters to $q$. If the number of parameters in $\mathbf{X}$ is logarithmic in the total number of parameter probabilities, i.e., $|\mathbf{X}| \leq p(\log |\Gamma|)$ for any polynomial $p$, then the problem is in $\mathsf{P}^\mathsf{PP}$, since we can try all combinations of parameter settings to 0 or 1 in polynomial time, using a PP-oracle.

If both the number of CPTs containing one or more parameters in the set $\mathbf{X}$ is bounded by a factor $k$ (independent of the number of total number of parameter probabilities), and the indegree of the corresponding nodes is bounded, then PARAMETER TUNING is PP-complete. Hardness follows immediately since PARAMETER TUNING has INFERENCE as a trivial special case (for zero parameters). We will prove membership of PP for this problem for a single parameter in a root node and show that the result also holds for a $k$-bounded number of CPTs with $m$ parents. Similar observations can be made for the other tuning problems defined in Section 4.

**Theorem 4.** PARAMETER TUNING *is* PP-*complete if the number of CPTs containing parameters and the indegree of the corresponding nodes are bounded.*

*Proof.* First let us assume $k = 1$, i.e., all $n$ parameters are taken from the CPT of a single node $V$. Furthermore, let us assume for now that $V$ is a root node. To solve PARAMETER TUNING, we need to decide whether $\Pr(C = c) \geq q$ for a particular combination of values of the parameters in $\mathbf{X}$. Conditioning on $V$ gives us $\sum_i \Pr(C = c \mid V = v_i) \cdot \Pr(V = v_i)$. Since $\sum_i \Pr(V = v_i) = 1$, $\Pr(C = c)$ is maximal for $\Pr(V = v_i) = 1$ for a particular $v_i$. Thus, if we want to decide whether $\Pr(C = c) \geq q$ for a particular combination of values of the parameters, then it suffices to determine whether this is the case when we set $\Pr(V = v_i) = 1$ for a particular parameter $v_i$.[2]

Using this observation, we will construct a Probabilistic Turing Machine $\mathcal{M}$ by combining several machines accepting INFERENCE instances. At its first branching step, $\mathcal{M}$ either accepts with probability $\frac{1}{2}$, or runs, with probability $\frac{1}{2n}$, one of $n$ Probabilistic Turing Machines $\mathcal{M}_i (i = 1, \ldots, n)$, which on input $\mathbf{B}_{\phi,i}$ (with $\Pr(V = v_i) = 1$) and $q$ accept if and only if $\Pr(C = c) = q$. If any $\mathcal{M}_i$ accepts, then $\mathcal{M}$ accepts. The majority of computation paths of $\mathcal{M}$ accepts if and only if the PARAMETER TUNING instance is satisfiable. If $V$ is *not* a root node, then we must branch over each parent configuration. For $k$ CPTs with at most $n$ parameters in each CPT and $m$ incoming arcs, we need to construct a combined Probabilistic Turing Machine consisting of $O(n^{m^k})$ Probabilistic Turing Machines accepting instances of INFERENCE. For bounded $m$ and $k$, this is a polynomial number of machines and thus computation takes polynomial time. Thus, PARAMETER TUNING is in PP for a bounded number of CPTs containing parameters and a bounded indegree of the corresponding nodes $m$ and $k$. □

## 7 Conclusion

In this paper, we have addressed the computational complexity of several variants of parameter tuning. Existing algorithms for sensitivity analysis and parameter tuning (see e.g. [2]) have a running time, exponential in both the treewidth of the graph and in the number of parameters varied. We have shown that parameter tuning is indeed hard, even if the network has a restricted polytree and if the number of parameters is bounded. We conclude, that PARAMETER TUNING is tractable only if *both* constraints are met, i.e., if probabilistic inference is easy *and* the number of parameters involved is bounded.

---

[2]If the number of parameters subject to tuning does not constitute all parameter probabilities in the CPT, then we need to test whether $\Pr(C = c) \geq q$ when all parameters have the value 0 as well.

**Acknowledgments**

This research has been (partly) supported by the Netherlands Organisation for Scientific Research (NWO). The authors wish to thank Hans Bodlaender and Gerard Tel for their insightful comments on earlier drafts of this paper. We wish to thank the anonymous reviewers for their thoughtful remarks.